**Remote Sensing Image Super-resolution and Object Detection: Benchmark and State of the Art**

Yi Wang [a, *], Syed Muhammad Arsalan Bashir [a, b, *], Mahrukh Khan [c], Qudrat Ullah [d], Rui Wang [a], Yilin Song [a], Zhe Guo [a], and Yilong Niu [e]

[a] School of Electronics and Information, Northwestern Polytechnical University, Xi'an, Shaanxi, China

[b] Quality Assurance, Pakistan Space and Upper Atmosphere Research Commission, Karachi, Sindh, Pakistan

[c] Department of Computer Science, National University of Computer and Emerging Sciences, Karachi, Sindh, Pakistan

[d] School of Automation, Northwestern Polytechnical University, Xi'an, Shaanxi, China

[e] School of Marine Science and Technology, Northwestern Polytechnical University, Xi'an, Shaanxi, China

* These authors contributed equally to this work.

**Emails:** Y Wang: wangyi79@nwpu.edu.cn; S M A Bashir: smarsalan@mail.nwpu.edu.cn, smab1176@yahoo.com; M Khan: mahrukhkhan0992@gmail.com; Q Ullah: qudratqaisrani@mail.nwpu.edu.cn; R Wang: wangrui2020@mail.nwpu.edu.cn; Y Song: 916135273@qq.com; Z Guo: guozhe@nwpu.edu.cn; Y Niu: yilong_niu@nwpu.edu.cn;

## Abstract

For the past two decades, there have been significant efforts to develop methods for object detection in Remote Sensing (RS) images. In most cases, the datasets for small object detection in remote sensing images are inadequate. Many researchers used scene classification datasets for object detection, which has its limitations; for example, the large-sized objects outnumber the small objects in object categories. Thus, they lack diversity; this further affects the detection performance of small object detectors in RS images. This paper reviews current datasets and object detection methods (deep learning-based) for remote sensing images. We also propose a large-scale, publicly available benchmark Remote Sensing Super-resolution Object Detection (RSSOD) dataset. The RSSOD dataset consists of 1,759 hand-annotated images with 22,091 instances of very high resolution (VHR) images with a spatial resolution of ~0.05 m. There are five classes with varying frequencies of labels per class; the images are annotated in YOLO and COCO format. The image patches are extracted from satellite images, including real image distortions such as tangential scale distortion and skew distortion. The proposed RSSOD dataset will help researchers benchmark the state-of-the-art object detection methods across various classes, especially for small objects using image super-resolution. We also propose a novel Multi-class Cyclic super-resolution Generative adversarial network with Residual feature aggregation (MCGR) and auxiliary YOLOv5 detector to benchmark image super-resolution-based object detection and compare with the existing state-of-the-art methods based on image super-resolution (SR). The proposed MCGR achieved state-of-the-art performance for image SR with an improvement of 1.2dB PSNR compared to the current state-of-the-art NLSN method. MCGR achieved best object detection mAPs of 0.758, 0.881, 0.841, and 0.983, respectively, for five-class, four-class, two-class, and single classes, respectively surpassing the performance of the state-of-the-art object detectors YOLOv5, EfficientDet, Faster RCNN, SSD, and RetinaNet.
Keywords: Remote sensing benchmark, multiclass GAN, object detection in remote sensing, small object detection, deep learning object detection, MCGR

## Introduction

Object detection and recognition have been a core problem in computer vision, and it aims to localize the objects within an image. As far as remote sensing is concerned in the past decade, object detection is a crucial task that is closely related to various applications, for instance, geographic mapping of resources, crop harvest analysis, disaster management, traffic planning, and navigation (Bai et al., 2021; Bashir & Wang, 2021b, 2021a; Kussul et al., 2017; Pi et al., 2020; Wei & Liu, 2021). Due to its comprehensive coverage, remote sensing images have various objects of interest ranging from small to large-sized objects. Therefore, object detection in RS images is a challenging task due to its multiscale nature, but recent advances in deep learning have provided a breakthrough in object detection and localization.

The deep learning-based methods are data-hungry, and their detection efficiency depends on the quality and quantity of the input data. A comprehensive and challenging dataset will help the advancement of object detection methods, for instance, the ImageNet (Deng et al., 2009) and MSCOCO (T. Y. Lin et al., 2014) datasets have been the standard for natural scene classification and object detection since their introduction, and most of the state-of-the-art methods are evaluated using these datasets. At the same time, the UC Merced (Y. Yang & Newsam, 2010) and NWPU-RESISC45 (Cheng et al., 2017a) datasets promote the progress in scene classification while the ISPRS Vaihingen (Rottensteiner et al., 2012) and 38-Cloud dataset (Mohajerani et al., 2018) set the path for the development of deep learning models for semantic segmentation of remote sensing images. The recent datasets like DOTA (Xia et al., 2018) and DIOR (Li et al., 2020) dealt with generic objects in remote sensing. These datasets, among many others, facilitated researchers to develop novel data-driven methods for object detection and recognition in the last decade (Alganci et al., 2020; Bashir & Wang, 2021b; Kousik et al., 2021; Stuparu et al., 2020; Yao et al., 2021).

In object detection for optical RS images, the object of interest (in most cases) occupies few pixels, for instance, a vehicle (with $4\times1.5$ $m^2$ footprint) in a very high resolution (VHR) image with a ground sampling distance (GSD) of 0.5 m will only occupy a pixel grid $8\times3$ (a total of 24 pixels). For small objects, such as vehicles, a few pixels represent the whole object; thus, identification and detection become challenging. A very high resolution (VHR) satellite image with a GSD of 0.25 m will have a 96 pixel ($16\times6$) areal grid for a vehicle with a dimension of $4\times1.5$ $m^2$. Thus, a recent development in remote sensing object detection is to perform object detection tasks in low-resolution (LR) images is to utilize the concept of image super-resolution, which increases the spatial resolution of the LR images as done by Courtrai et al. (Courtrai et al., 2020) and Bashir et al. (Bashir et al., 2021). There was a need to benchmark such methods that utilize low-resolution images to perform object detection on a challenging dataset. Thus, the Remote Sensing image Super-resolution Object Detection (RSSOD) dataset will facilitate those researchers to benchmark the methods and detection tasks performed on high-resolution (HR) images.

The main contributions of this paper are as follows:

1. A large-scale public dataset, RSSOD[1] for small-object detection tasks in the urban

[1] RSSOD will be publicly available at: DOI: 10.17632/b268jv86tf.1

environment (sample images shown in Figure 1). With its 22,091 hand-annotated instances and five classes, this dataset will provide the researchers with a challenging detection due to its geographic information and orientation of the annotations.

2. We provide the performance benchmarking on the proposed RSSOD dataset with the current state-of-the-art.
3. We also propose a novel Multi-class Cyclic super-resolution Generative adversarial network with Residual feature aggregation (MCGR) for object detection in low-resolution images for scale factors 2 and 4.

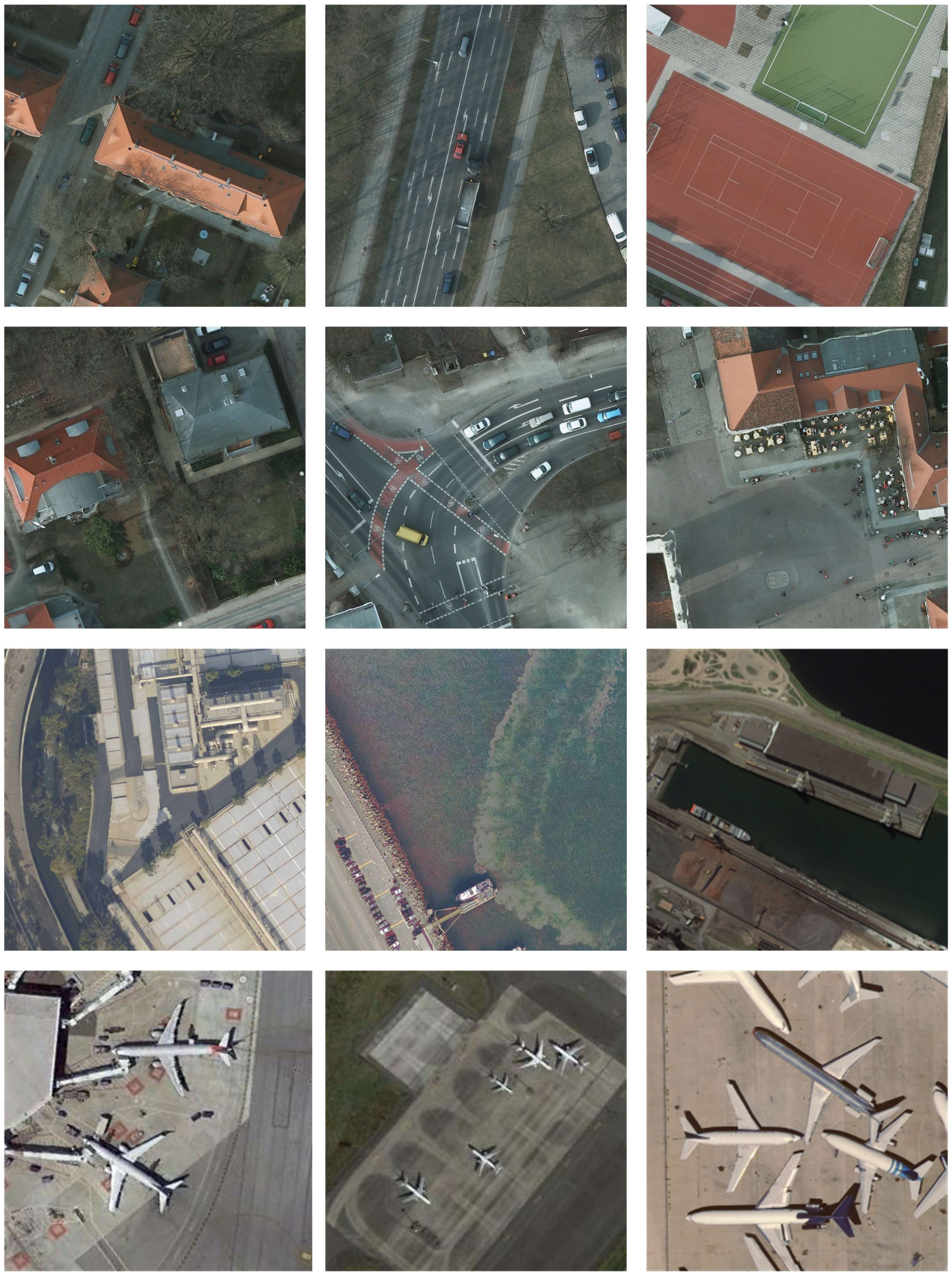

Figure 1. Visualization of the RSSOD dataset

The paper is organized as follows; Section 2 reviews previous datasets and methods for object detection in remote sensing images. In Section 3, we describe the proposed RSSOD dataset, while in Section 4, we describe the novel MCGR for object detection in LR images, evaluation metrics, and implementation details. Section 5 describes the benchmarking results on the RSSOD dataset for MCGR and the existing small-object detectors using state-of-the-art image SR methods and object detectors. Finally, the conclusion and future direction are discussed in Section 6.

## 2. A Review of Remote Sensing Object Detection Datasets and Methods

In the past decade, the remote sensing-based object detection dataset and methods have been explored by researchers., this is mainly because of the advancements in the satellite sensor design that high-quality VHR images are now available for deep learning. In this section, we briefly review the current datasets and methods for object detection in remote sensing.

### 2.1. Object Detection Datasets for RS Images

Since the images in remote sensing are a top view, and the object size is diverse due to changes in sensor design in the satellite payload, the object instances often have a high deviation in size and orientation. Some leading datasets include NWPU-RESISC45, NWPU VHR-10 (Cheng et al., 2016), DIOR, DOTA. In this section, we briefly introduce the current datasets in remote sensing object detection and recognition.

1) TAS Dataset (Heitz & Koller, 2008): The TAS dataset was designed to detect vehicles in aerial images as it contained 30 images with a total of 1319 cars in arbitrary orientation. The drawbacks are low resolution and dark shadows projected by the nearby building and trees.

2) UC Merced Land-Use Dataset (Y. Yang & Newsam, 2010): This is a 21-class scene classification dataset with 100 images measuring 256×256 per class. The images were extracted from USGS National Map Urban Area Imagery, and the spatial resolution is 0.3048 m. UC

Merced is one of the most used datasets in remote sensing scene classification, and it has overlapping classes, such as sparsely residential, high-density residential, and medium-density residential areas, making this dataset richer and more challenging. Some of the remote sensing image classification studies include LGFBOVW classifier (Zhu et al., 2016), LASC-CNN classifier (Yuan et al., 2018), Hybrid Satellite Image Classification System (A. et al., 2018), Structured Metric Learning (Gong et al., 2019), mcODM classifier (Z. Zhang et al., 2021), Feature Engineering-based Classifier (Rasche, 2021).

3) NWPU Datasets: The initial remote sensing image classification dataset proposed by researchers at NWPU was termed NWPU-RESISC45, which had a total of 45 classes with 700 images per class. The spatial resolution of the dataset was ~0.2 to 30 m, while the image size was fixed to 256×256. Due to its size (total of 31,500 images), this dataset is widely used for scene classification; however, the spatial resolution of most images is low, so another dataset with VHR images termed NWPU VHR-10 (Cheng et al., 2014) which included 715 images with a spatial resolution of 2 m and 85 images with a spatial resolution of 8cm. There were ten classes in the proposed dataset, and it also included negative examples. However, the drawback was that it included some small-sized objects which were not labeled as their size was only a few pixels (for example, vehicle class). Unlike NWPU-RESISC45, this dataset was an object detection dataset, and the horizontal boundary box (HBB) annotation was used in this dataset. The researchers frequently used these datasets for remote sensing scene classification and object recognition tasks (Cheng et al., 2016; Dong et al., 2019; Rasche, 2021; C. Wang et al., 2019; Xu et al., 2020; Xue et al., 2020).

4) VEDAI (Razakarivony & Jurie, 2016): This dataset by Razakarivony et al. was introduced for multiclass vehicle detection as it contained a total of 3,640 vehicle instances which included a total of 9 classes, i.e., boat, car, camping car, plane, pick-up, tractor, truck, van, and the other category. This dataset included a total of 1210 images with a spatial resolution of 12.5 cm and a size of 1024×1024. Researchers frequently use this dataset to benchmark object detection in remote sensing (Ju et al., 2021; Lu et al., 2018; Sakla et al., 2017; Vasavi et al., 2021). Another vehicle class dataset was DLR MVDA (K. Liu & Mattyus, 2015) which used the DLR 3K camera system to take VHR images at an altitude of 1000m and a spatial resolution of 13cm. DLR MVDA used oriented bounding boxes (OBB) for annotation with multi-direction boxes to indicate the direction of the vehicle.

5) RSC11 Dataset (Zhao et al., 2016): This dataset was also extracted from Google Earth, and it included 11 scene classes, including some very similar class scenes, which made classification difficult for this dataset. The spatial resolution of this dataset was 0.2 m and a total of 1232 images with an image size of 512×512.

6) ISPRS Potsdam (Rottensteiner et al., 2012): This dataset is a semantic segmentation dataset by the International Society for Photogrammetry and Remote Sensing. The dataset contains 38 VHR images of size 6000×6000 and a spatial resolution of 5 cm. This dataset has been widely used (Bashir & Wang, 2021b; Bokhovkin & Burnaev, 2019; Chen et al., 2018; Courtrai et al., 2020) for semantic segmentation and object detection tasks, especially in the urban environment, as the dataset includes six classes namely impervious surface, building, low vegetation, tree, car, and clutter.

7) RSOD Dataset (Xiao et al., 2015): Xiao et al. collected a total of 976 images with a spatial resolution of 0.5 – 2 m. This dataset has four classes, namely, oil tanks, airplanes, overpasses, and playgrounds.

8) DOTA Dataset: This dataset by Xia et al. is a large object detection dataset that has a total of 15 object classes, including large-sized objects (like bridge, harbor, basketball court) and small-sized objects (like a small vehicle). A total of 2,806 images with varying resolutions were included in the dataset with more than 188k object instances. Due to significant image size and spatial resolution variations, the objects have varying scales, orientations, and shapes, making the detection task more challenging.

9) DIOR Dataset: The DIOR dataset by Li et al. contains 23k images with more than 192k instances and 20 object categories (~1200 images per class). HBB based annotations were used in this dataset, while a later version DIOR-R was released with updated OBB annotations for OBB object detection tasks.

Table 1. A Comparison of RSSOD with other remote sensing datasets.

| **Dataset** | **Classes** | **Images** | **Instances** | **Avg. image size** | **Spatial Resolution (m)** | **Annotation style** | **Year** |
|---|---|---|---|---|---|---|---|
| TAS | 1 | 30 | 1319 | 636 | ~0.2 | HBB | 2008 |
| UC Merced Land-Use | 21 | 2100 | - | 256 | 0.3048 | Scene Class | 2010 |
| NWPU VHR-10 | 10 | 800 | 3775 | ~1000 | 0.08 - 2 | HBB | 2014 |
| VEDAI | 9 | 1210 | 3640 | 1024 | 0.125 | OBB | 2015 |
| DLR MVDA | 2 | 20 | 14235 | 5616 | 0.13 | OBB | 2015 |
| RSC11 | 11 | 1232 | - | 512 | 0.2 | Scene Class | 2016 |
| ISPRS Potsdam | 6 | 38 | - | 6000 | 0.05 | Semantic Segmentation | 2016 |
| RSOD | 4 | 976 | 6950 | ~1000 | 0.5 - 2 | HBB | 2017 |
| NWPU-RESISC45 | 45 | 31500 | - | 256 | 0.2 - 30 | Scene Class | 2017 |
| DOTA | 15 | 2806 | 188282 | ~800 to 4000 | ~0.4 | OBB | 2017 |
| DIOR | 20 | 23463 | 192472 | 800 | 0.5 - 30 | HBB+OBB | 2018 |
| RSSOD (ours) | 5 | 1759 | 22091 | 1000 | ~0.05 | HBB | 2021 |

HBB: horizontal boundary box; OBB: oriented boundary box

Table 1 depicts the comparison of the proposed RSSOD dataset with existing datasets. RSSOD dataset has the highest spatial resolution compared to other object detection datasets, and the dataset has high-resolution images, i.e., ~1000×1000 pixels, emphasizing small objects.

## 2.2. Object Detection Methods for RS Images

The object detection methods are mainly of two kinds, i.e., generic object detectors that perform the detection as an end-to-end learning task or SR-based methods that use a prior network to enhance the image quality before object detection. This section reviews the current generic object detectors, including some state-of-the-art small object detectors and image SR-based object detectors.

### *2.2.1. Universal Object Detectors*

In recent years, deep learning-based methods have been successfully implemented in classification, detection, and recognition tasks (Bao et al., 2020; Bashir & Wang, 2021a; Z. Lin et al., 2019; Wei & Liu, 2021). Among these tasks, object detection is one of the most prominent areas that gets expanded every year, where many object detectors are being developed and released. The detection task is either single-stagged or dual-stage; in the latter, the region proposals are generated, followed by the boundary box classification task.

One of the earliest 2-staged detectors was based on region-based convolutional neural networks (RCNN) (Girshick et al., 2014), which employed a selective search algorithm (Uijlings et al., 2013) for region proposals which were passed to CNN for generation of feature vectors for each proposed region. In R-CNN, the final classification task was performed by Support Vector Machines (SVM). R-CNNs were superseded by Fast RCNN (Girshick, 2015), which introduced two main changes in the initial RCNN; the first change was the introduction of shared feature maps among the region of interest. The second change was to replace the SVM with a fully connected (FC) neural network for object classification and boundary box regression; this allowed end-to-end training, leading to real-time object detection. Another modification of RCNN, Faster RCNN (Ren et al., 2015), where the authors introduced a cost-free Region Proposal Network (RPN), which resulted in 5fps with state-of-the-art results on Pascal VOC datasets.

A single-stage detector initially was used to achieve high fps at the cost of detection accuracy; for instance, the You-Only-Look-Once (YOLO) model (Redmon et al., 2016) achieved 45 fps with a mean Average Precision (mAP) of 63.4% on Pascal VOC 2007 and 2012 while Faster RCNN achieved mAP of 70.0 with 0.5 FPS. Thus, the era of real-time object detectors began, which opened the realm of real-time object detectors on video streams. Another one-stage detector is the Single-Shot Detector (SSD) (W. Liu et al., 2016) which does object localization and classification in a single pass by eliminating the region proposal phase. Boundary boxes having fixed sizes are used for object detection, and the final detection is performed using non-max suppression. By incorporating focal loss in SSD, RetinaNet (T. Y. Lin et al., 2017) was able to surpass the accuracies of all state-of-the-art two-stage detectors while keeping the speed of one-stage.

Redmon et al. proposed the initial three versions of YOLO; YOLO-v2 (Redmon & Farhadi, 2017) and YOLO-v3 (Redmon & Farhadi, 2018). These new detectors incorporated deeper backend CNNs, residual blocks, and skip connections, leading to a state-of-the-art performance with a 3.8× higher speed than RetinaNet. In 2020, YOLOv4 was proposed by Bochkovskiy et al. (Bochkovskiy et al., 2020), which utilized a new backend network called CSPDarknet53 with spatial attention and Mish-activation to improve the mAP and fps (on MS COCO Dataset) by 10% and 12%, respectively compared to YOLOv3. Within one month of YOLOv4, Jocher released the YOLOv5 (Jocher et al., 2020) version, which achieved state-of-the-art performance mAP and fps. The YOLOv5x6 model reported the highest reported mAP of 55.4% on MS COCO Val 2017 dataset with an inference time of 19.4ms per image and is currently ranked top among state-of-the-art object detectors; a comparison of various models with EfficientDet (Tan et al., 2020) is shown in Figure 2.

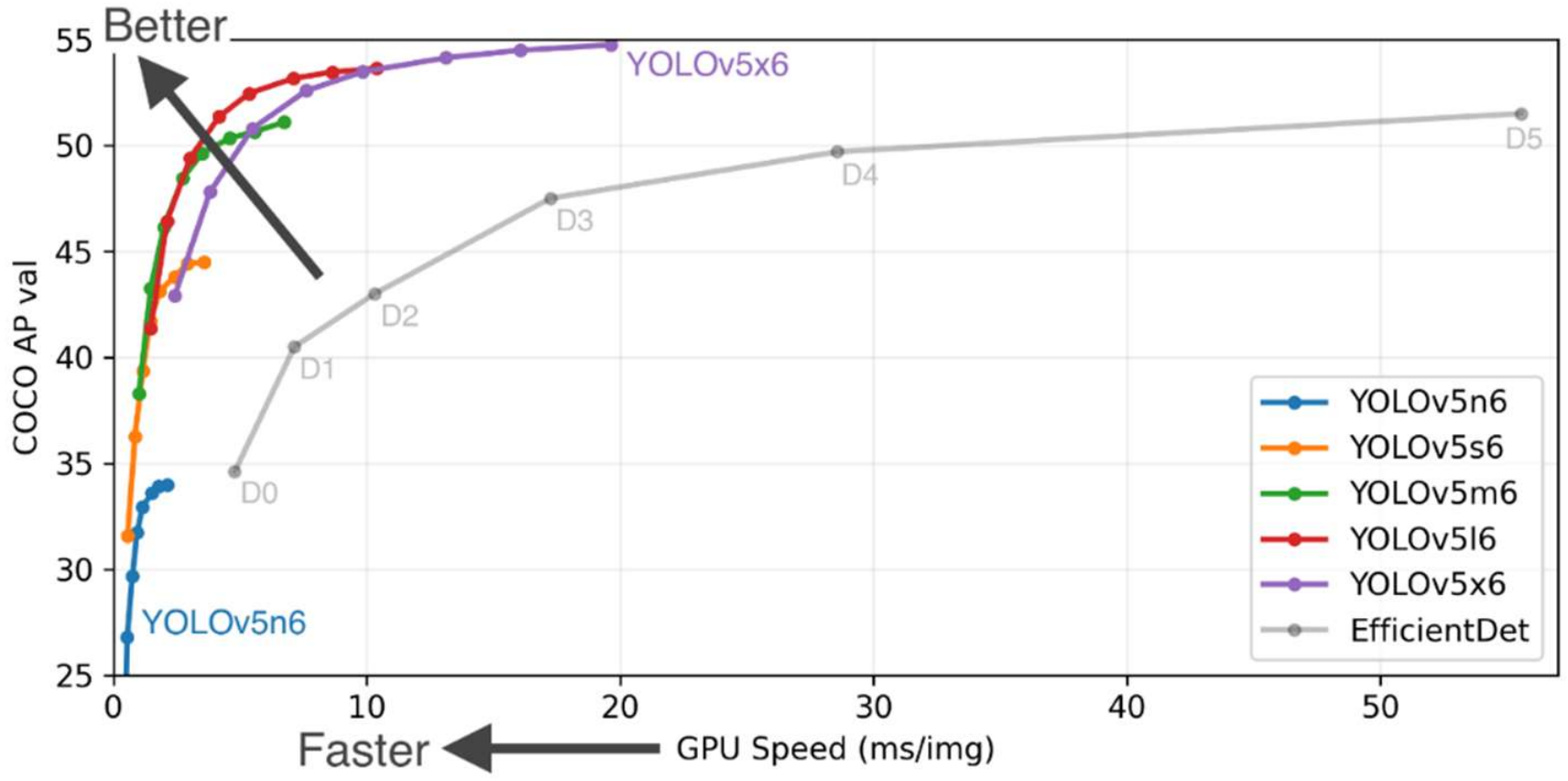

Figure 2. Detection accuracy and inference time comparison of YOLOv5 models with EfficientDet (Jocher et al., 2020)

#### *2.2.2. Detectors for Small-sized Objects in RS Images*

Here we discuss some state-of-the-art image SR methods for LR to HR image SR that are further used in small object detection for RS images by various researchers in the recent past.

##### *2.2.2.1. State-of-the-art Image SR Methods*

Here we discuss some of the state-of-the-art methods for image SR. In 2017, Lim et al. introduced Enhanced Deep SR network (EDSR) (Lim et al., 2017). In recent times (Rabbi et al., 2020; Shermeyer & Van Etten, 2019; Wei & Liu, 2021), the EDSR has been an integral part of small object detection in remote sensing images, but the reported detection mAP of such detectors was low.

Wang et al. proposed improvements in the existing SR Generative Adversarial Network (GAN) architecture and proposed ESRGAN for realistic image SR (X. Wang et al., 2019). Wang et al. introduced Residual-in-Residual Dense Block (RRDB) along with adversarial and perceptual loss. Further improvements were reported by Wang et al. as they trained the ESRGAN with pure synthetic data using a high-order degradation modeling that was close to the real-world degradations (X. Wang et al., 2021).

Liang et al. proposed SwinIR, which used transformers to address image SR by incorporating three parts within the SwinIR transformer: shallow feature extraction, deep feature extraction, and high-quality image reconstruction using the Residual Swin Transformer Blocks (RSTB) (Liang et al., 2021). SwinIR reported state-of-the-art restful on DIV2K(Agustsson & Timofte, 2017) and Flickr2K datasets.

In image SR, the overall performance of the model is dependent upon the degradation model used to generate LR images; most models consider additional factors such as blur; still, the real-world degradations are diverse; to address this issue, Zhang et al. proposed a more practical degradation model in BSRGAN (K. Zhang et al., 2021). BSRGAN degradation model generates LR images with random blue shuffle, downsampling, and noise degradations, making the degradation more realistic.

In DRN (Guo et al., 2020), a dual regression network was proposed to learn the mapping of LR to HR and the corresponding degradation mapping function, i.e., HR to LR, which learns the down-sampling kernel. This method achieved state-of-the-art performance in terms of PSNR and the number of parameters compared to methods like EDSR, RCAN (Y. Zhang et al., 2018), and RRDB.

Mei et al. proposed a Non-local Sparse Attention (NLSA) (Mei et al., 2021) method for image SR; the non-local attention is done with spherical hashing that divides the input into hash buckets of relevant features, thereby the noisy and locations with less information are not given attention during the learning process. The network was named as NLSN for non-local sparse network, and it achieved state-of-the-art performance Set14 (Zeyde et al., 2012), B100 (Martin et al., 2001), Urban100 (Huang et al., 2015), Manga109 (Fujimoto et al., 2016) datasets.

##### *2.2.2.2. Image SR-based Small-Object Detectors*

The object's size is a critical parameter in object detection tasks, especially the overall inference image. Current state-of-the-art detectors work efficiently on medium and large size objects, but when it comes to small-sized objects (i.e., dimension in a few pixels or occupying less than 5% of the overall image size, the performance of generic object detectors degrades exponentially (Bashir & Wang, 2021b; Courtrai et al., 2020). Due to small size, the features of small objects are indistinguishable from the features of other classes, which leads to detection inaccuracies. One way to improve the detection performance is to perform data augmentation by oversampling the small objects of interest by simply copying the small objects (Kisantal et al., 2019). The augmentation technique increases the possibility of overlap with the prediction and ground truth, thereby increasing the prediction accuracy. However, this technique will decrease the detection accuracies of other object classes due to overlap during the augmentation process; even with 0 overlap with other objects, the negative samples are decreased due to the pasting of objects over the background, which may increase in false-positive rate during detection tasks. Another approach to small object detection is to train the network for small and large objects on multiple resolutions, as done by Park et al. (Park et al., 2010).

For small object detection, an auxiliary image SR network could increase the spatial resolution of the dataset before the actual detection task, as performed by Courtrai et al. and Bashir et al. (Bashir & Wang, 2021b; Courtrai et al., 2020). In recent years the image SR networks have achieved significant results for scale factors of 2× and 4×. Some deep learning-based methods to generate HR images from LR include single-image SR (SISR) (Hui et al., 2021), which performs the SR task by taking a single image as input. For a detailed review of all image super-resolution methods, including conventional and deep learning-based methods, we encourage readers to read review papers by Yang et al. (W. Yang et al., 2019) and Bashir et al. (Bashir et al., 2021).

A two-stage detector with image SR using GAN and object detection using SSD was proposed by Ferdous et al. (Ferdous et al., 2019), while Zhang et al. used weakly supervised learning to learn object detection in RS images using a pseudo-label generation method (L. Zhang & Ma, 2021). Alternatively, Rabbi et al. combined ESRGAN and Edge-Enhanced GAN (EEGAN) (Jiang et al., 2019) to develop an end-to-end small object detection network that used Faster RCNN and SSD for object detection (Rabbi et al., 2020).

## 3. The Proposed RSSOD Dataset

In this section, we present the RSSOD dataset along with the information on data collection, class selection, annotation method, data split, image sizes, spatial resolution, and object orientations.

### 3.1. Image Collection

Initially, we reviewed remote sensing scene classification and object detection datasets (see Table 1) and realized the lack of a high-resolution urban object detection dataset. We collected the data from already published datasets for other tasks such as scene classification. The urban images were extracted from the ISPRS Potsdam semantic segmentation dataset, where we extracted 1000×1000 pixel-sized patches from 38 VHR image tiles with an overlap of 100 pixels. For reproducibility, we used the same names from the original dataset and used an additional two-digit to denote the patch number; a total of 36 patches were extracted from single images, i.e., a total of 1368 images. Overall image selection of the RSSOD dataset is shown in Table 2. The average image resolution is 856×853.

Table 2. Details of image selection for RSSOD dataset.

| **Dataset Name** | **VHR Images** | **Extracted Patches** | **Size** | **Spatial Resolution (m)** |
|---|---|---|---|---|
| ISPRS Potsdam | 38 | 1368 | 1000×1000 | 0.05 |
| UC Merced Land-Use | - | 99 | 256×256 | 0.3048 |
| NWPU-RESISC45 | - | 101 | 256×256 | 0.8 |
| Draper Satellite Image Chronology (Draper, 2016) | 9 | 11 | 1000×1000 | 0.2 |
| Ship Images from Google | - | 180 | ~421×388.5 | 0.8 |
| | | Total: 1759 | Avg: 856×853 | |

Avg: average

The images were taken from multiple sensors, and the orientations of objects are random, and the location of the objects are random, as seen in Figure 3a. The majority of the objects have small sizes, as shown in Figure 3b.

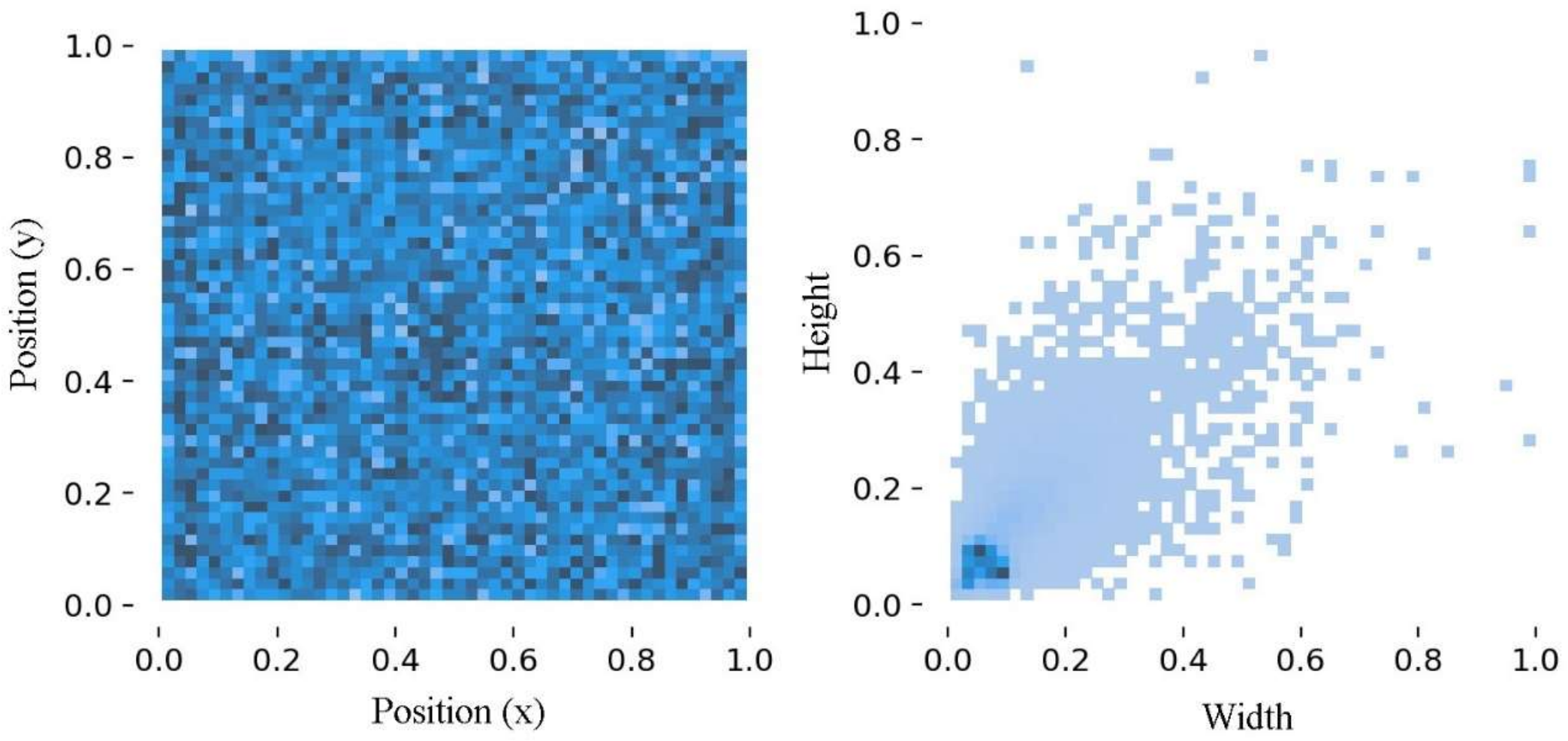

(a) (b)

Figure 3. Frequency plot for object location (left) and size (right). The dimensions are normalized on a 0 – 1 scale.

### 3.2. Class Selection and Data Split

In remote sensing, the object of interest is of utmost importance; in most of the object detection datasets, the common classes include a vehicle (including ground vehicles, ships, and airplanes), tree, building, impervious surface, and low vegetation (Cheng et al., 2017b; Rottensteiner et al., 2012). Considering the object size, we omitted the building class as most of the images are from an urban setting and the buildings occupy most of the image space; therefore, the final considered classes include vehicle, tree, airplane, ship, and low vegetation. The details of class instances are shown in Table 3.

Table 3. Class distribution for RSSOD dataset.

| Class Name | Total Instances |
|---|---|
| Vehicle | 10603 |
| Tree | 6775 |
| Airplane | 441 |
| Ship | 277 |
| Low Vegetation | 3995 |
| Total | 22091 |

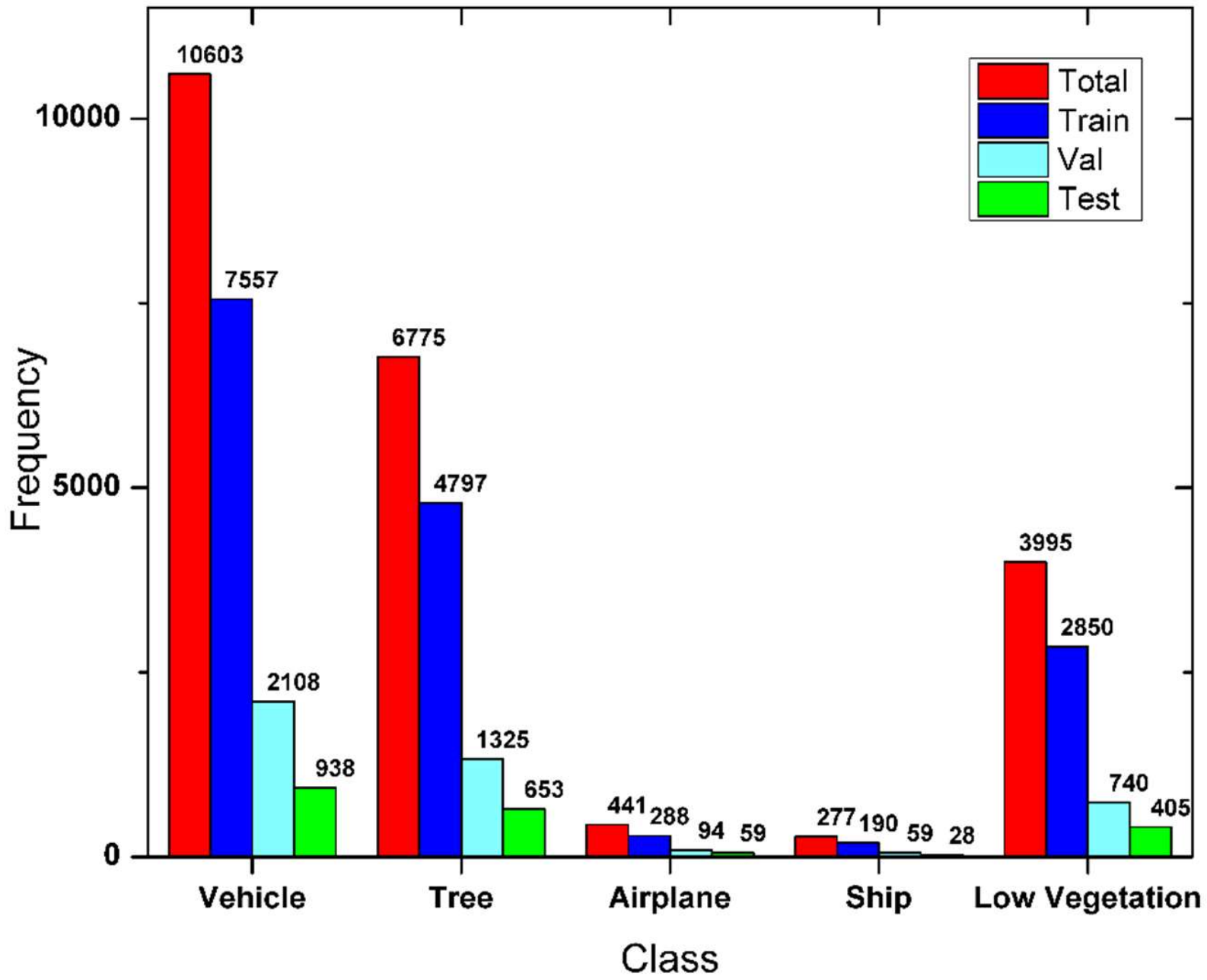

Figure 4. Class frequencies for the train-validation-test subsets.

The dataset was split as per the conventional 70:20:10 ratio for training, validation, and testing subsets. The overall instance distribution for the training, validation and testing images is depicted in Figure 4. The training, validation, and test examples are 1232, 351, 176, respectively.

### 3.3. Annotation Method

For annotation, we selected horizontal boundary box-based annotation, and a modified form of a Python-based open-source labeling tool OpenLabeling (Cartucho et al., 2018), was used for labeling the images. All the objects were manually labeled using the YOLO format using one point and width, the height of the boundary box (BB), i.e., <class identifier, *x*, *y*, *w*, *h*>; where (*x*, *y*) denote the center of the BB while (*w*, *h)* denote the width and height respectively.

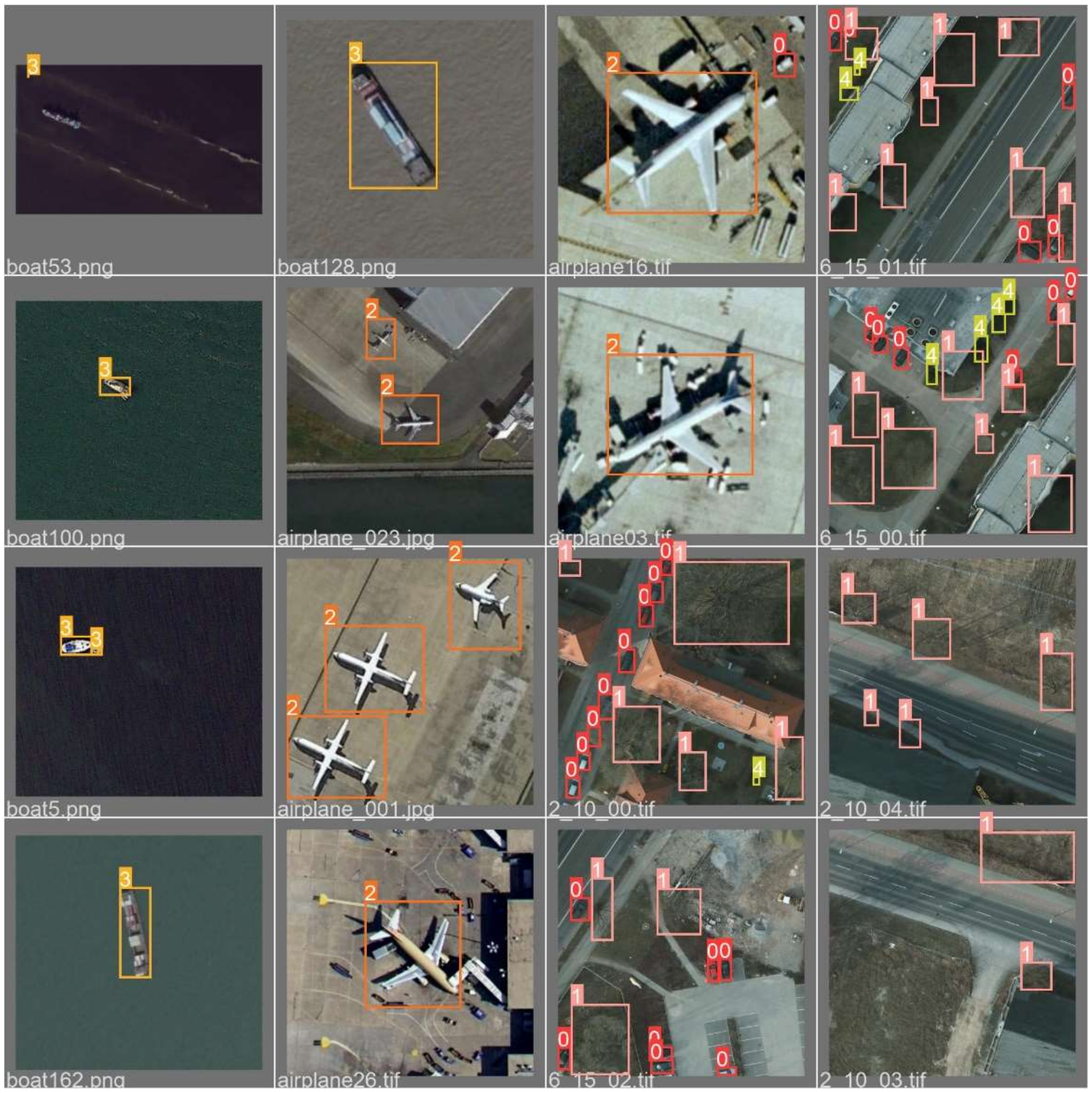

Figure 5. RSSOD annotations where the object classes are shown on a scale of 0 – 4.

The annotations were refined by manually reviewing in three rounds, and any human error in labeling was addressed during the review process. Furthermore, the annotations are grouped into four sets, including five, four, two, and one class. This was done to facilitate object detection on various classes and see the performance of the detectors for the subset of classes. Some of the annotated images from RSSOD are shown in Figure 5.

## 4. Small-object Detection using Multiclass Cyclic GAN with RFA

In this section, we proposed a novel Multiclass Cyclic GAN with residual feature aggregation (RFA) for object detection on the RSSOD dataset. The proposed network is based on two tasks, i.e., image SR and Object detection. We first introduce the proposed image SR network.

### 4.1. MCGR for Image SR-based Object Detection

We first replaced the conventional residual blocks in image SR as proposed in SRResNet (Ledig et al., 2017) and EDSR with RFA-based blocks (see Figure 6), which aggregate the features from all residual blocks using 1×1 convolution layer as proposed by Bashir et al. (Bashir & Wang, 2021b). This feature concatenation enhanced the performance of the network and thereby increased the quality of the SR image.

Furthermore, we use a modified form of SRGAN, which includes RFA-based residual blocks and the cyclic network based on Wasserstein GANs (Arjovsky et al., 2017) to train the network using L1 and L2 loss functions; the network diagram of the proposed cyclic GAN is shown in Figure 7.

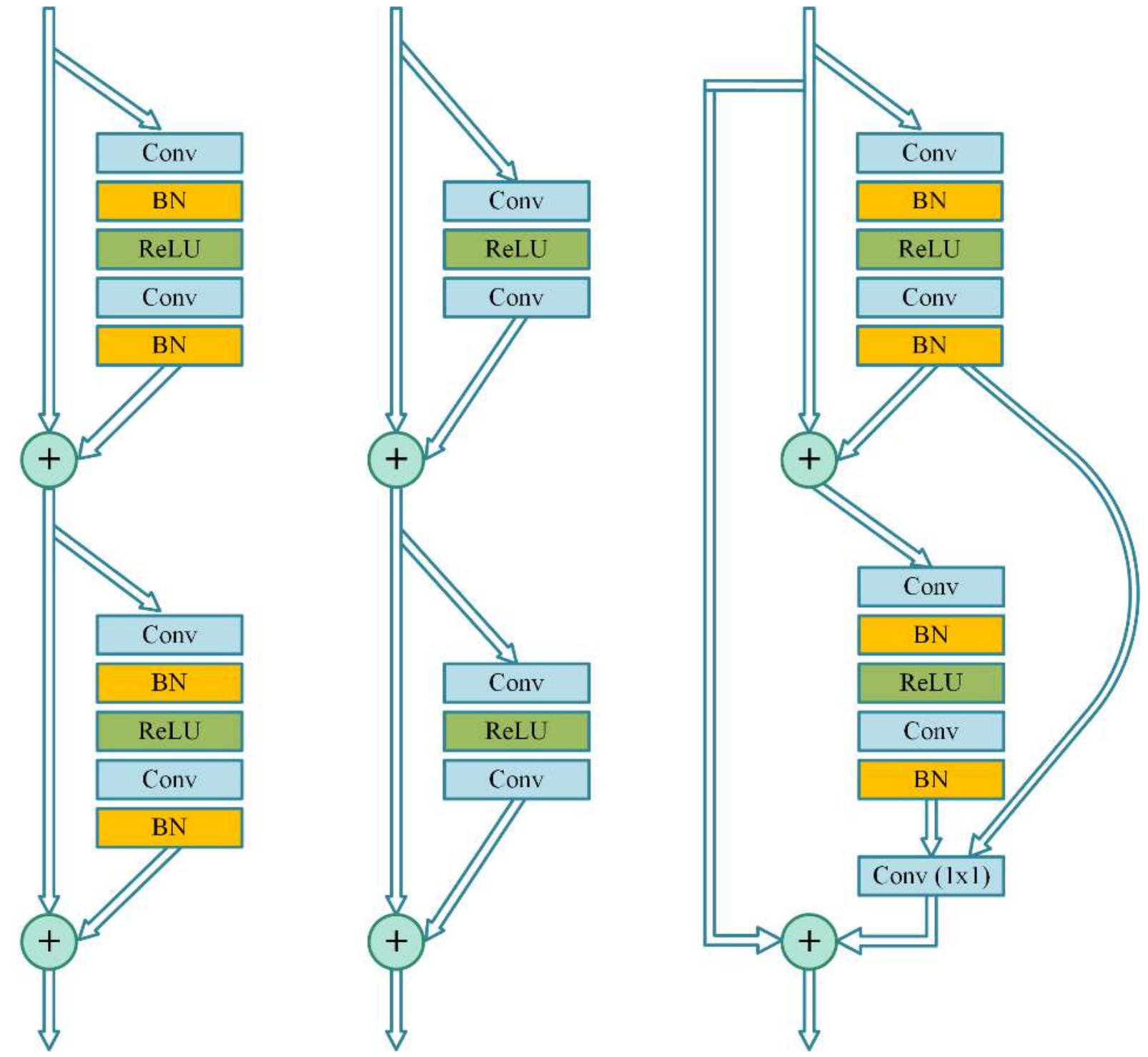

(a) SRResNet (b) EDSR (c) Proposed

Figure 6. Proposed Residual Feature Aggregation as compared to other residual blocks.

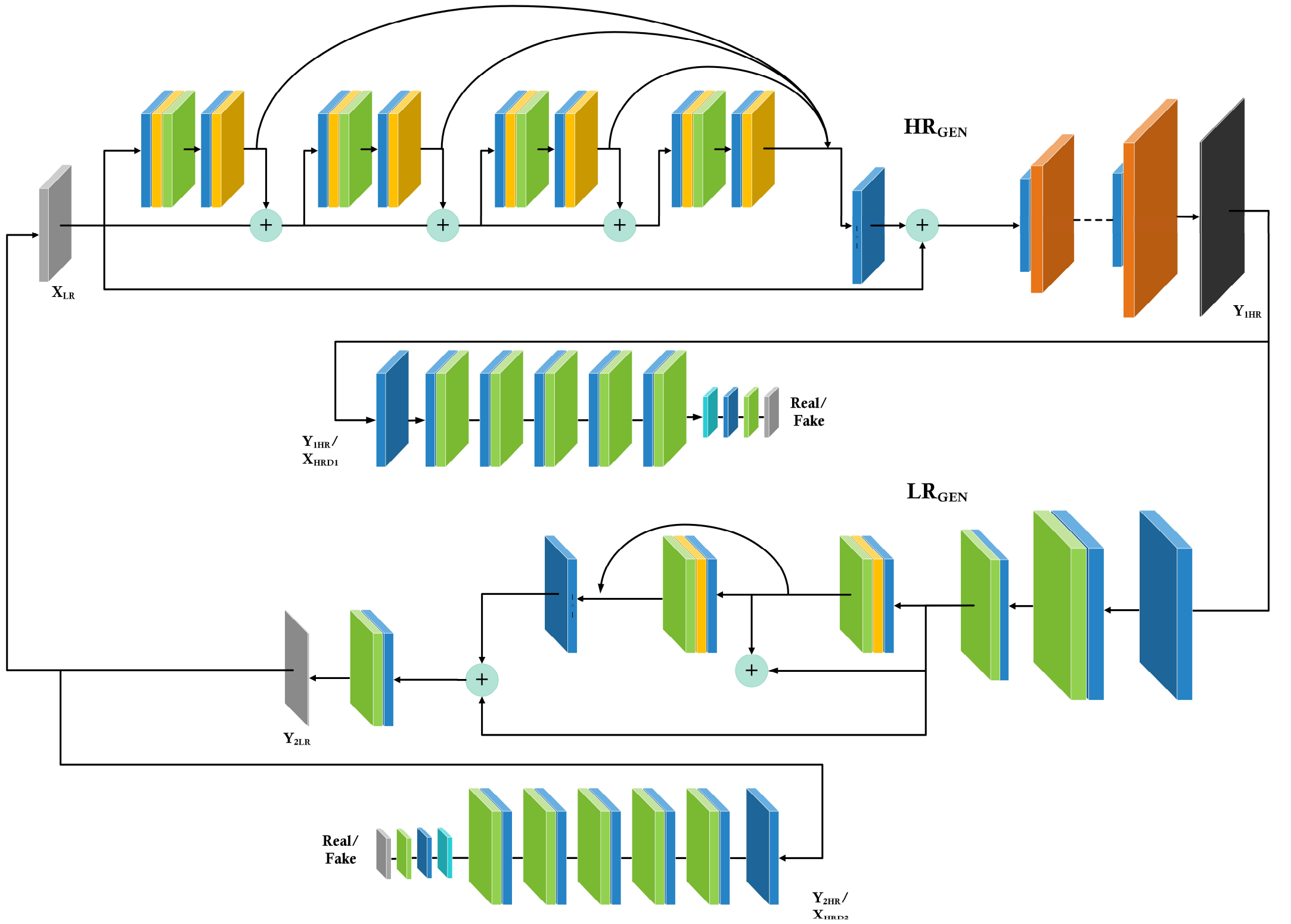

Figure 7. The cyclic GAN with RFA; the top part is the HR generator and its discriminator, while the bottom is the LR generator and its discriminator.

The color codes are similar to Figure 6, with light blue color for the reduction layer and brown for the pixel rearrangement layer. The cyclic GAN (as shown in Figure 7) uses a second GAN, $LR_{GEN,}$ to generate the LR image from the HR image generated by $HR_{GEN}$, the overall loss function of this cyclic model is shown in Equation 1.

$$\begin{aligned} \mathcal{L}(SRCGAN) &= \mathcal{L}_{L1}\left(HR_{GEN}\left(I_{LR}\right), I_{HR}\right) + \mathcal{L}_{MSE}\left(HR_{GEN}\left(LR_{GEN}\left(I_{HR}\right)\right), I_{HR}\right) + \\ &\mathcal{L}_{L1}\left(LR_{GEN}\left(I_{HR}\right), I_{LR}\right) + \mathcal{L}_{MSE}\left(LR_{GEN}\left(HR_{GEN}\left(I_{LR}\right)\right), I_{LR}\right) \end{aligned} \quad (1)$$

where, $I_{HR}$ and $I_{LR}$ are HR, and LR images; $HR_{GEN}$ and $LR_{GEN}$ are the corresponding HR and LR generators, as shown in Figure 7.

By comparing the LR-$LR_{GEN,}$ we ensure that the generated LR image is similar to the actual LR, and thereby the generated HR image by the network will also be similar to the actual HR image. This cyclic approach ensures that the two GANs minimize the overall loss by evaluating the output generated by one another. Furthermore, a detection network is used at the end, a YOLOv5 detector, as shown in Figure 8.

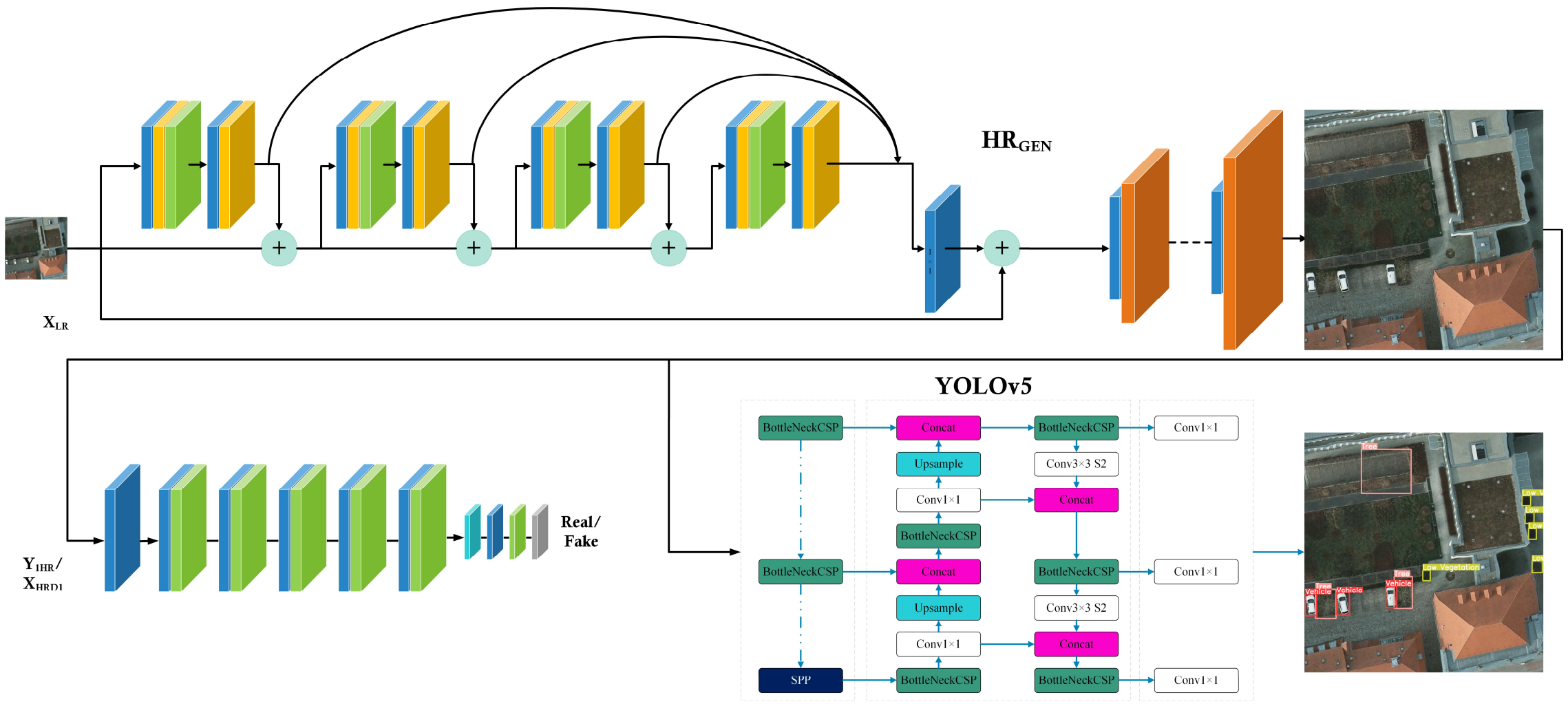

Figure 8. The network architecture for Multiclass Cyclic GAN with RFA (MCGR).

The three networks are shown in Figure 8 have different loss functions, and the total loss function is a weighted sum of the loss functions of Generator, Discriminator, and Detector.

The loss function of the generator network is given in Equation (2):

$$\mathcal{L}(HR_{GEN}) = \frac{1}{N}\sum_{i=1}^{N}\left\| HR_{GEN}\left(I_{LR}^{i}\right) - I_{HR}^{i}\right\|_1 \tag{2}$$

where, $\mathcal{L}(HR_{GEN})$ is loss of generator network, $N$ is the total number of samples, $HR_{GEN}\left(I_{LR}^{i}\right)$ is the $i-th$ generated HR image, and $I_{HR}^{i}$ is the *i-th* ground truth HR image.

Using the improved WGAN with YOLOv5, a 48 block RFA based generator with a block size of 64 × 64 and kernel of 3 × 3; a high gradient penalty coefficient was used in the discriminator network and the discriminator loss $\mathcal{L}(Dis)$ is given in Equation (3).

$$\mathcal{L}(Dis) = \sum_{I_{HR}\sim\mathbb{P}_{HR}}\left[Dis\left(I_{HR}\right)\right] - \sum_{I_{SR}\sim\mathbb{P}_{SR}}\left[Dis\left(I_{SR}\right)\right] + \lambda\sum_{I_{ran}\sim\mathbb{P}_{ran}}\left[\left(\left\|\Delta Dis\left(I_{ran}\right)\right\|_2 - 1\right)^2\right] \tag{3}$$

Where $I_{HR}$, $I_{SR,}$ and $I_{ran,}$ are the HR, SR, and a uniformly sampled randomly selected image from $\{I_{HR}, I_{SR}\}$. The probability density distributions of HR, SR, and random images were $P_{HR}$, $P_{SR,}$ and $P_{ran,}$ respectively. The gradient penalty coefficient $\lambda$ was assigned a high value of 10.

The final YOLOv5 network has a boundary box loss function, as shown in Equation (4).

$$\mathcal{L}(Y5) = \sum_{i=0}^{Det^2}\sum_{j=0}^{Anc^2}\left[\left(x - x'\right)^2 + \left(y - y'\right)^2 + \left(w - w'\right)^2 + \left(h - h'\right)^2\right] \tag{4}$$

where *Det* and *Anc* represent detection grid and the number of anchor points while the coordinates of the HR and SR predicted BB are represented by *(x, y, w, h)* and *(x', y', w', h')*. *(x, y)* and *(x', y')* represents the center point of the BB while *(w, h)* and *(w', h')* represent the width and height of the ground truth and the predicted BB, respectively. The network in Figure 8 is trained using a total loss function $\mathcal{L}(Tot)$ as shown in Equation (5).

$$\mathcal{L}(T) = \mu_1\mathcal{L}(HR_{GEN}) + \mu_2\mathcal{L}(Dis) + \mu_3\mathcal{L}(Y5) \tag{5}$$

The weight coefficients $\mu_1$, $\mu_2$, and $\mu_{3,}$ are 0.90, 10, and 0.10, respectively. This selection was used to normalize the errors generated by the three networks, thereby bringing stability to the training process. Hence, using the SR image generated by the generator, the object class identification and object localization are performed by the detection network.

### 4.2. Methods for Evaluation

The proposed method utilizes the concept of image SR to improve the image quality before the detection task. Therefore, we use image quality metrics such as PSNR, SSIM, and Task evaluation (object detection) to compare and contrast the performance of the proposed SR network with the existing state-of-the-art methods in Section 5.1.

For object detection, we measure the mAP@0.5 IoU for object detection across classes and compare it with the existing methods for object detection using YOLOv5 as an auxiliary detector. In recent times this metric has been used to evaluate the object detectors (Bashir & Wang, 2021b; Ferdous et al., 2019; W. Liu et al., 2016; Rabbi et al., 2020; Redmon & Farhadi, 2018, 2017). Since most objects are small in size, a low value of IoU and Confidence threshold was selected during the inference task.

### 4.3. Implementation Details

The proposed network was implemented using the PyTorch framework and Ubuntu 20.04 computer with Titan XP graphics processor by Nvidia. During the training phase, the coefficients of the total loss function $\mu_1$, $\mu_2$ and $\mu_3$ were 0.90, 10, and 0.10, respectively, and the network was trained for 100 epochs.

## 5. Benchmark Results of RSSOD

In this section, we share the benchmarking results on the RSSOD dataset using state-of-the-art image SR and object detection methods. The training was performed on the training set with validation of the validation set, while the final evaluations were based on the test set. We also discuss the benchmarking results to provide insights about the proposed MCGR network for object detection in remote sensing images using the proposed RSSOD dataset.

## 5.1. Image Quality Assessment

The SR results are evaluated using three primary quantitative Image Quality Assessment (IQA) metrics, i.e., MSE, PSNR, and SSIM. Furthermore, we also compare the SR image quality for recovery of complex textures between the proposed MCGR and the state-of-the-art NLSN method. The LR version is for a scale factor of 4; we compare the results of MCGR with HR image, Real-ESRGAN, SwinIR-L, BSRGAN, DRN, EDSR, NLSN, as shown in Figure 9 and 10.

As seen in Figure 9, the Real-ESRGAN, SwinIR-L, and BSRGAN do over smoothing; thereby, the ground texture information beneath the cars is not recovered. Also, DRN blends the information from the neighboring pixels, which results in low image quality. The EDSR and NLSN recover the high-level information within the image, but the texture information, including the high-frequency details, are recovered by the proposed MCGR network. Figure 10 depicts the results of image SR for the airplane class, which confirms the superior performance of the proposed MCGR network.

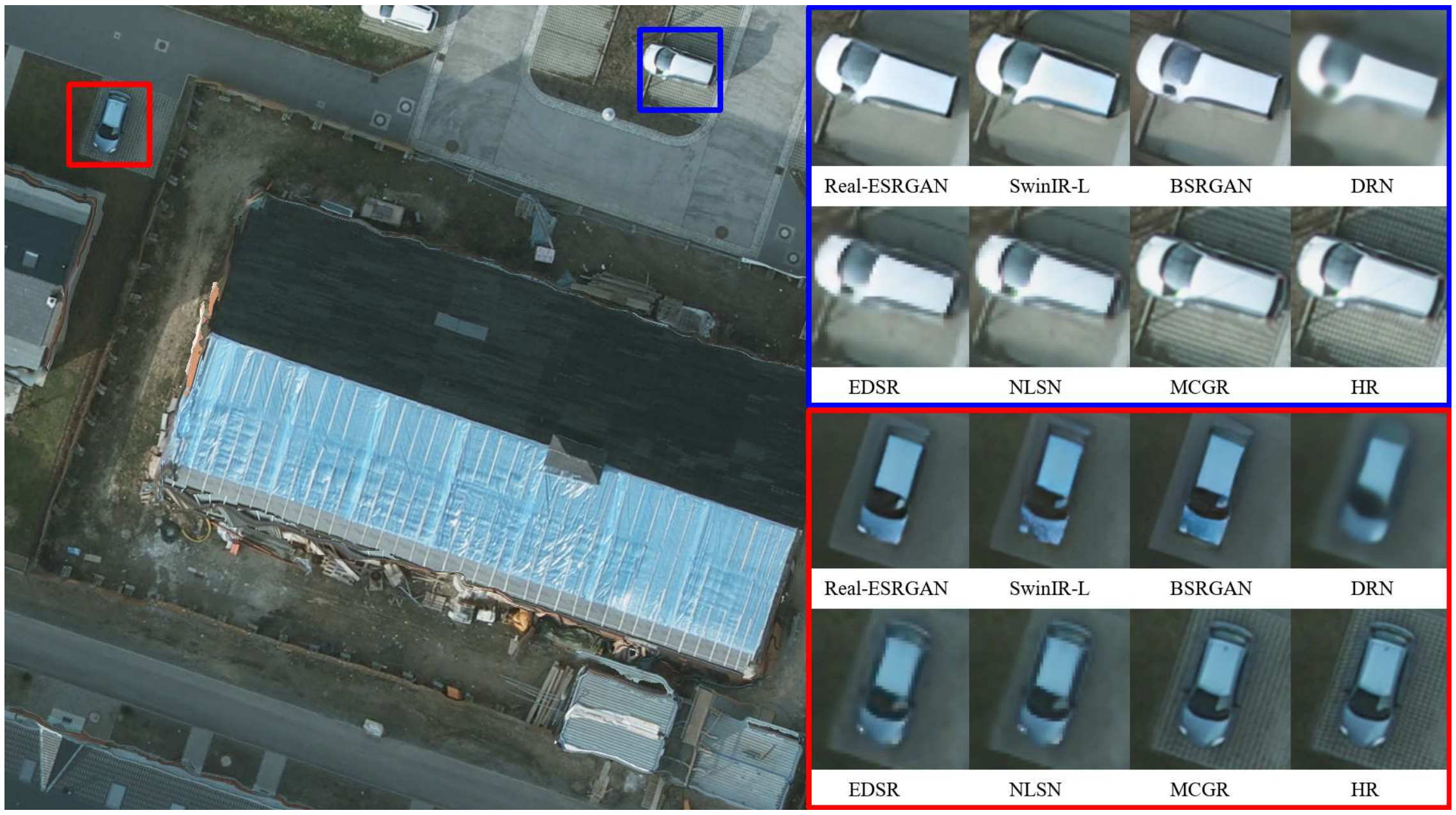

Figure 9. Image quality assessment of proposed MSCG with Real-ESRGAN, SwinIR-L, BSRGAN, DRN, EDSR, NLSN, and HR for a scale factor of 4 for object class vehicle.

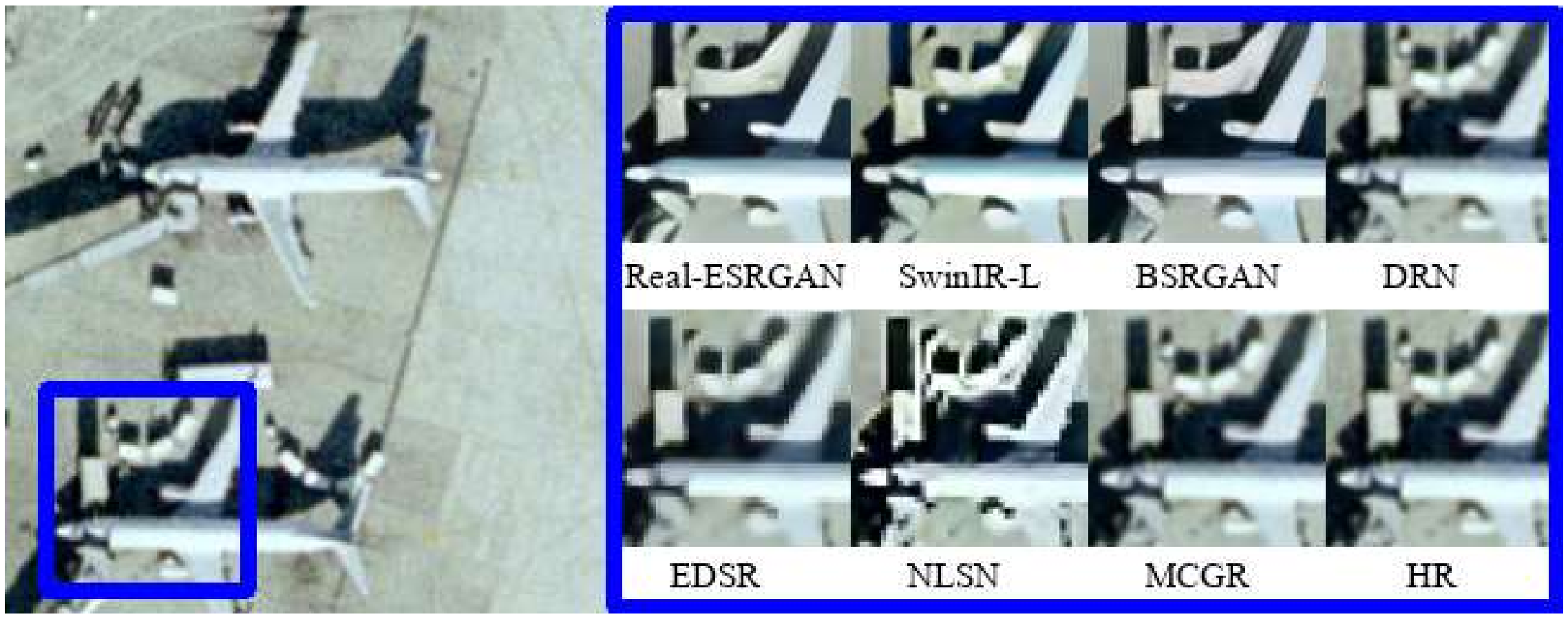

Figure 10. Image quality assessment of proposed MSCG with Real-ESRGAN, SwinIR-L, BSRGAN, DRN, EDSR, NLSN, and HR for a scale factor of 4 for object class airplane.

We also show the image SR results in comparison to the second-best performing method NLSN in Figure 11. Here the zoomed patches depict the proposed MCGR to recover real-world textures from the LR image using RFA-based cyclic GAN. The colors denote the object location while the left half four boxes are the results of MCGR, and to the right are the results of NLSN. The zoomed images in Figure 11 show the quality of recovered details for the two compared methods.

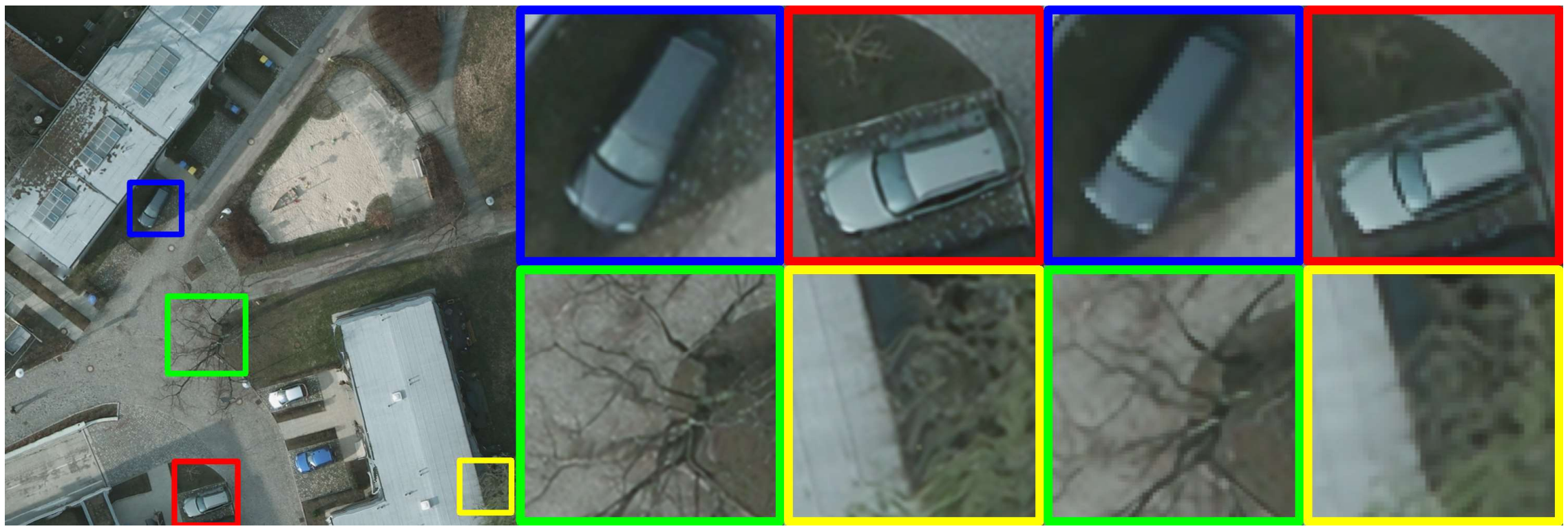

(a) MCGR (b) NLSN

Figure 11. Image quality assessment of proposed MSCG with second best method NLSN.

The summary of IQA metrics for the test set is shown in Table 4. The average PSNR for bicubic-interpolation was 30.23 dB, while the proposed MCGR achieved state-of-the-art performance in terms of MSE, PSNR, and SSIM. MCGR achieved the best PSNR of 34.68 dB, while NLSN, EDSR, and DRN achieved PSNR of 33.48 dB, 33.13 dB, and 32.69 dB, respectively. MCGR achieved the best MSE of 27.98, while NLSN, EDSR, and DRN achieved PSNR of 33.48 dB, 33.13 dB, and 32.69 dB, respectively. MCGR achieved PSNR of 34.68 dB, while NLSN, EDSR, and DRN achieved PSNR of 31.17, 37.64, and 44.89, respectively. The best result of 0.93 for SSIM was achieved by MCGR, while NLSN, EDSR, and DRN achieved SSIM of 0.88, 0.89, and 085, respectively.

Table 4. The performance of MCGR image SR on the Test Set for a scale factor of 4.

| **Method** | **MSE** | **PSNR** | **SSIM** |
|---|---|---|---|
| Bicubic | 79.99 | 30.23 | 0.79 |
| Real-ESRGAN | 72.54 | 30.81 | 0.81 |
| SwinIR-L | 65.34 | 30.90 | 0.76 |
| BSRGAN | 58.26 | 31.85 | 0.82 |
| DRN | 44.89 | 32.69 | 0.85 |
| EDSR | 37.64 | 33.13 | <u>0.89</u> |
| NLSN | <u>31.17</u> | <u>33.48</u> | 0.88 |
| MCGR | **27.98** | **34.68** | **0.93** |

Note: Best and second-best results are bold and underlined, respectively.

The proposed MCGR achieved the best results for recovering high-frequency information from the LR image, as seen in Figures 9 – 11 and Table 4. Here we also share the training performance using the precision/recall curves for the training of four separate networks with a scale factor of 4 for five-class, four-class, two-class, and one-class object detection. The training was conducted for 100 epochs, and the achieved mAP for detection and precision/recall curves are shown in Figure 12.

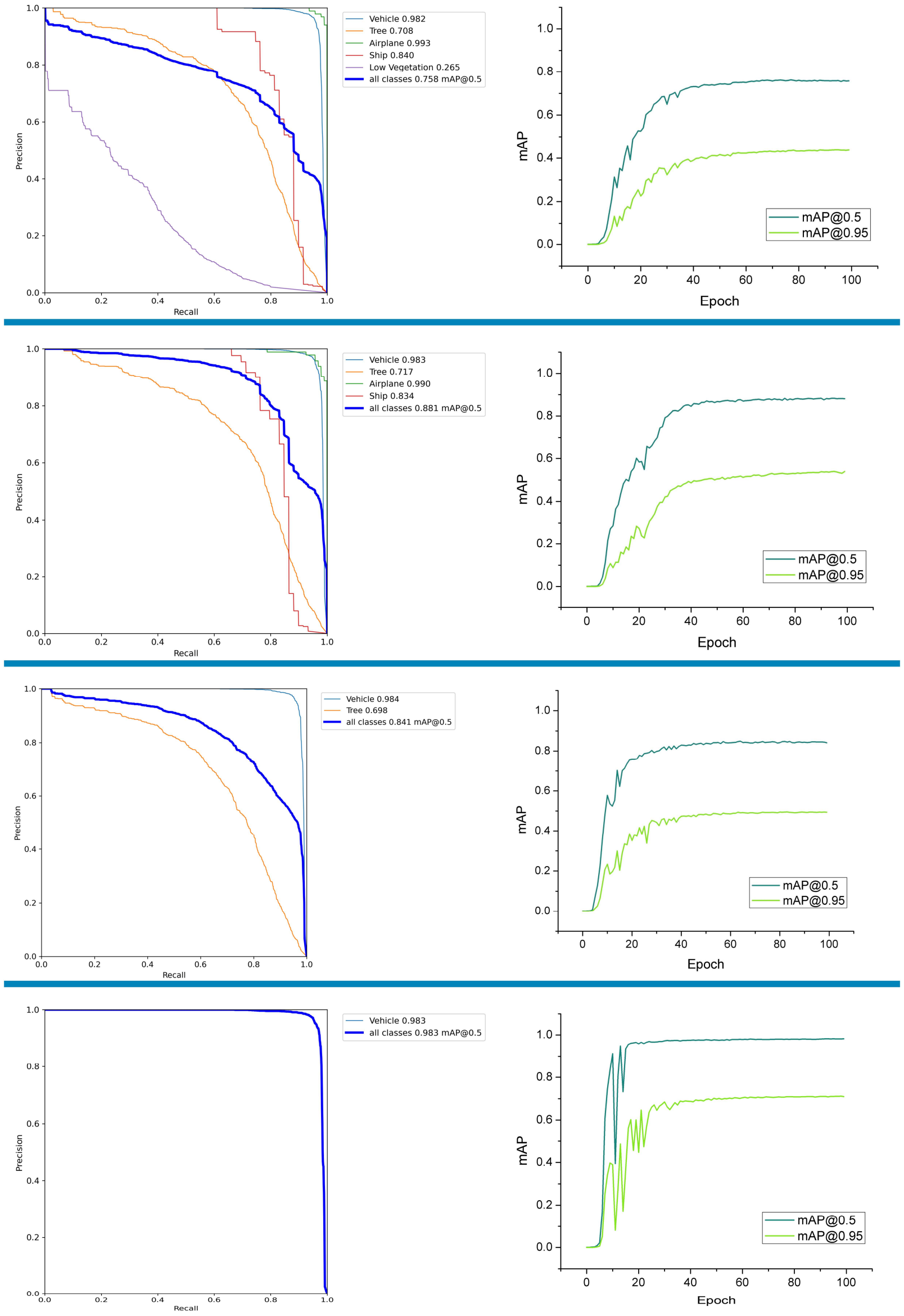

(a) (b)

Figure 12. Training performance for various classes in terms of mAP and Precision/recall curves. (From top to bottom, five, four, two and one class training results) (a) Precision/recall curves (b) mAP curves for IoU of 0.50 and 0.95

As evident from the training performance shown in Figure 12, the proposed MCGR converges very fast for one and two classes (i.e., under 25 epoch), while for four classes, the network achieves stable mAP after 40 epochs with a minimum mAP of 0.717 for tree class while the highest mAP of 0983 was achieved for vehicle class. For five classes, the learning becomes a little difficult as the low vegetation class is quite similar to the tree class, and therefore mAP of 0.265 was achieved at 0.5 IoU. The training mAP for the four models trained on five, four, two, and one class are 0.758, 0.881, 0.841, and 0.983, respectively.

The achieved mAP for various classes and detection results are further elaborated in the next section.

### 5.2. Benchmarking for Object Detectors

In this section, we discuss the results of object detection on the RSSOD dataset in comparison to the state-of-the-art detectors such as RetinaNet (T. Y. Lin et al., 2017), SSD (W. Liu et al., 2016), Faster RCNN (Ren et al., 2015), EfficientDet-D5 (Tan et al., 2020) and the proposed MCGR. To avoid bias during the test inference, we used the HR images from the RSSOD-train set to train the object detectors with validation of the validation set. The results for five-class object detection for an input size of 640×640 pixels on the RSSOD-test set are shown in Table 5. For HR-test images, the best detection mAP of 0.76 is reported by YOLOv5, while the second-best mAP of 0.746 is achieved by MCGR. It is also clear that as the scale factor (SF) increases, the detection performance of the generic detectors deteriorates as the object size shrinks; however, the MCGR achieved mAP of 0.731 and 0.711 for SF of 2 and 4, respectively.

Table 5. The impact of scale factor on the detection mAP with an IoU of 0.10 on a five-class test set.

| **Method** | **HR (mAP)** | **SF = 2 (mAP)** | **SF = 4 (mAP)** | **Time (ms)** |
|---|---|---|---|---|
| YOLOv5 | **0.76** | <u>0.68</u> | <u>0.58</u> | **4.33** |
| RetinaNet | 0.50 | 0.44 | 0.21 | 93.64 |
| SSD (VGG16) | 0.51 | 0.43 | 0.27 | 31.10 |
| Faster R-CNN | 0.68 | 0.53 | 0.44 | 88.14 |
| EfficientDet-D5 | 0.72 | 0.57 | 0.46 | 52.31 |
| MCGR | <u>0.746</u> | **0.731** | **0.711** | <u>16.61</u> |

SF: scale factor, best: bold, second-best: underlined

The proposed network achieves state-of-the-art performance on both HR and LR images with a relatively reasonable inference time per image (i.e., 16.61ms). Figure 13 shows the detection performance for a few images on the test set.

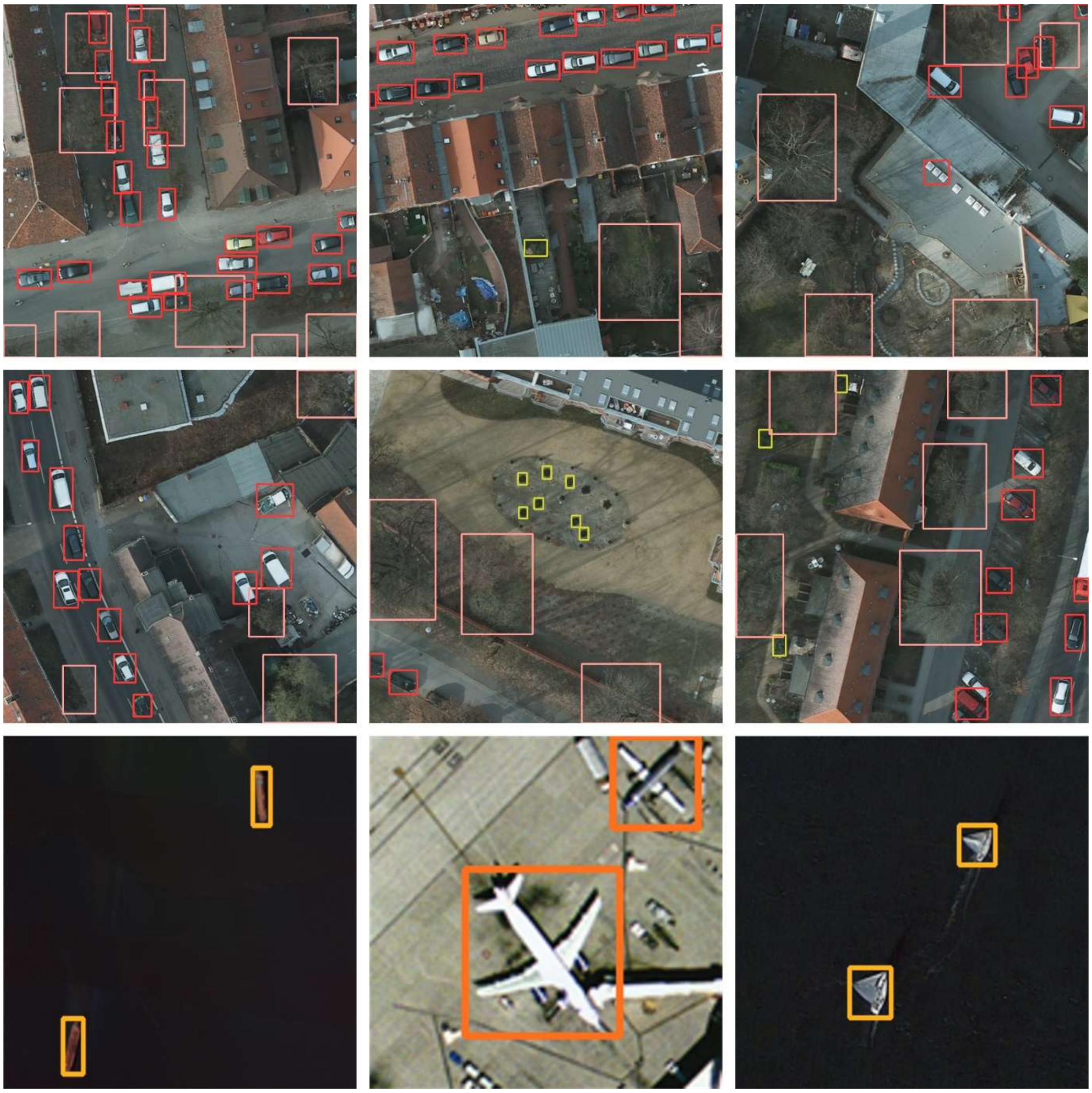

Figure 13. Results of object detection using MCGR. Red, pink, orange, bright yellow, and pear color represent vehicle, tree, airplane, ship, and low vegetation classes, respectively

The detection of small objects by MCGR using the image SR has significantly improved the object detection, however, the inference time was four times higher than the YOLOv5, but this was significantly better than the other reported methods, which shows the superior performance of the proposed MCGR network.

### 5.3. MCGR Performance on Independent Dataset

We also benchmark our MCGR network for an independent dataset DroneDeploy (Pilkington et al., 2019) and share the object detection results in Figure 14 for 1000×1000 pixels patches to validate the performance of the proposed MCGR.

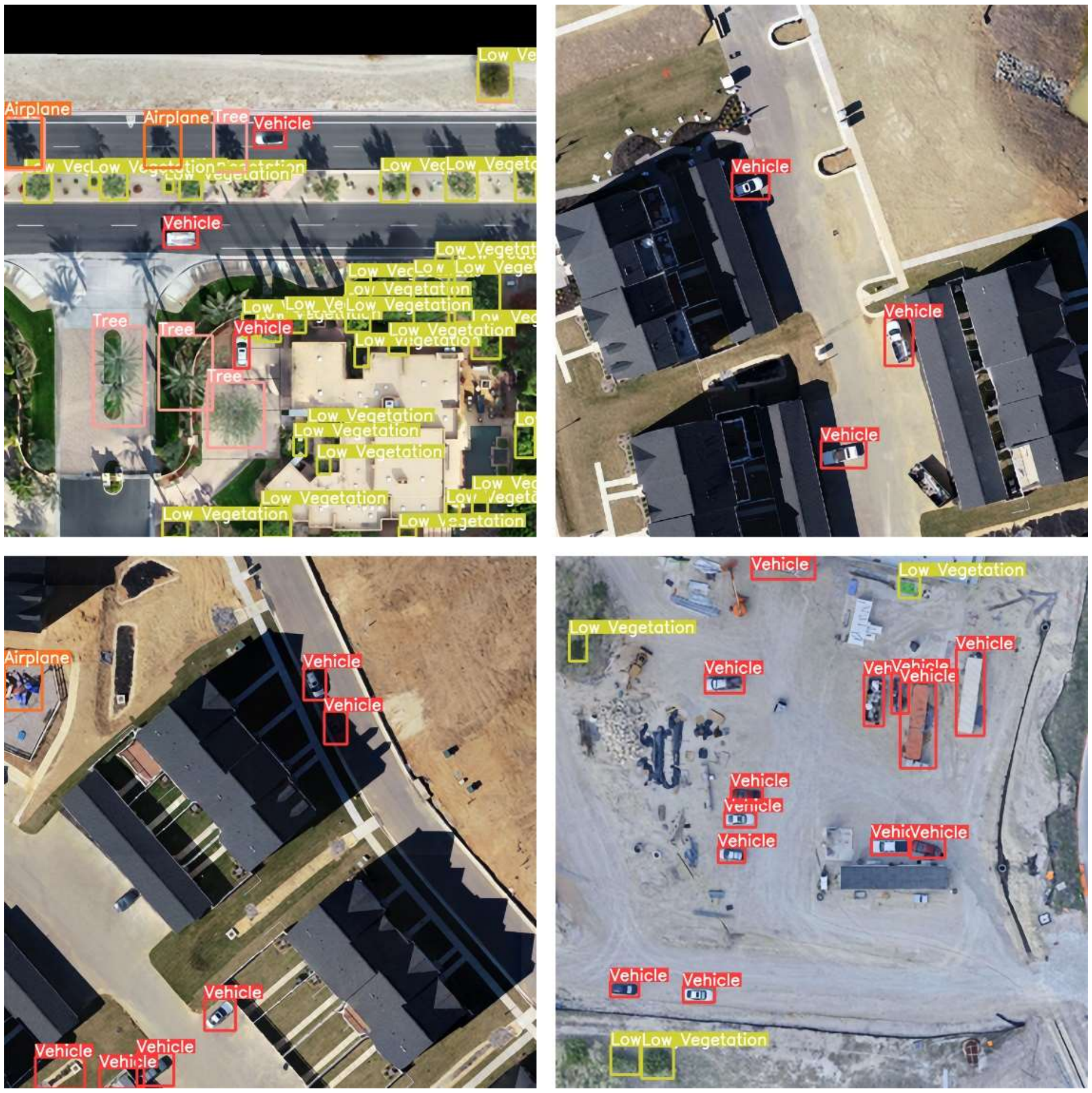

Figure 14. Results of MCGR for object detection on DroneDeploy Dataset.

We also tested MCGR for another aerial dataset by Gąsienica-Józkowy et al. called the Aerial dataset of Floating Objects (AFO), where the dataset contains tiny objects with six classes (Gąsienica-Józkowy et al., 2021). The proposed Ensembled method by Gąsienica-Józkowy achieved object detection mAP of 0.8216 at an IoU of 0.50 while our MCGR achieved state-of-the-art performance with mAP of 0.845, shown in Figure 15. The key challenge was to learn the features for small objects which occupied less than 1% of the total area of the images.

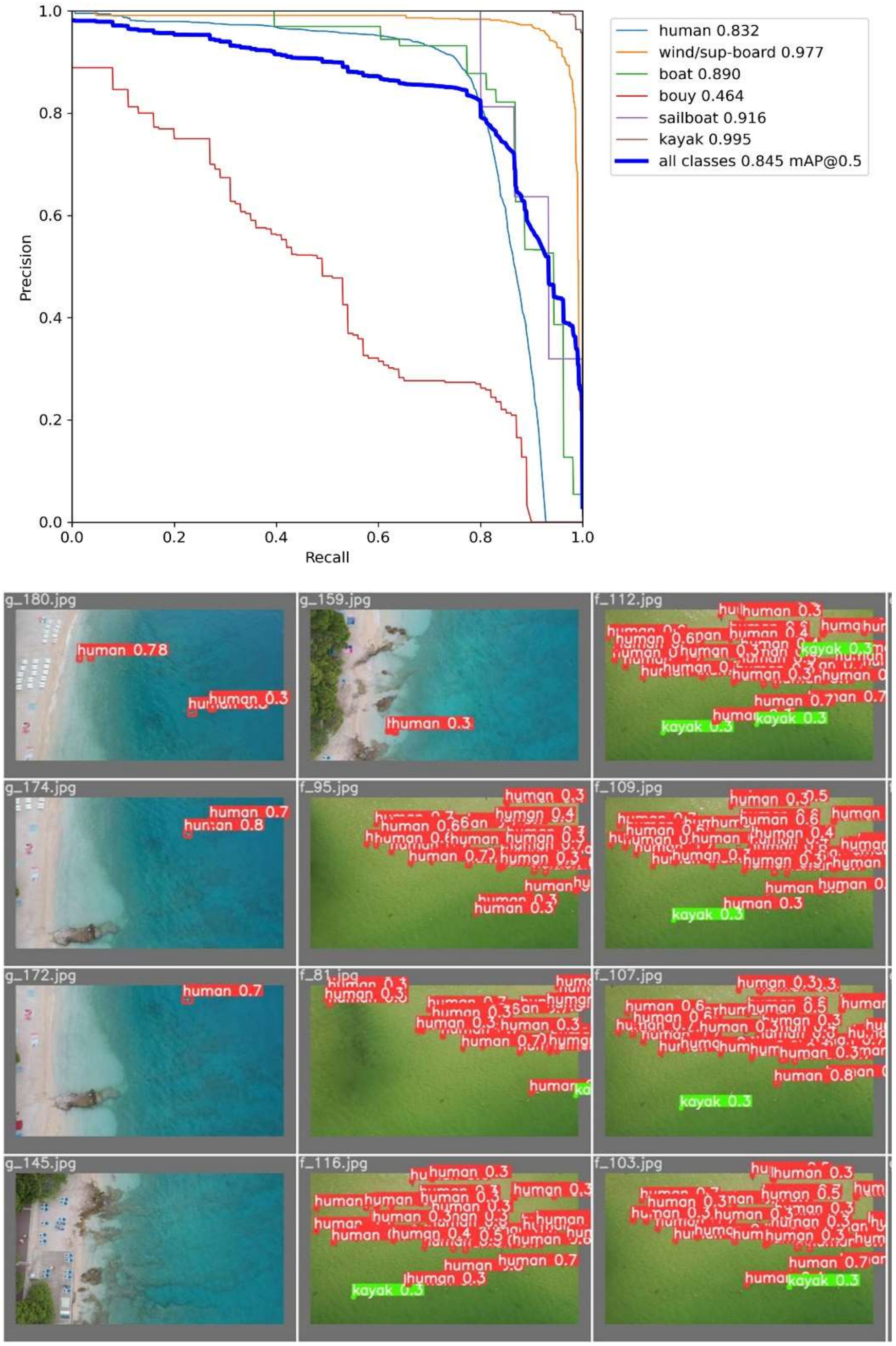

Figure 15. Results of MCGR transfer learning on AFO Dataset. Precision/recall curve on top while the detection results at the bottom.

## 6. Conclusion and Future Directions

The object localization and detection task in RS images is a topic of continuous research; thus, developing state-of-the-art object detectors for remote sensing of the environment is of utmost importance. In this paper, we proposed a new benchmark RSSOD dataset for remote sensing object detectors with a high overlap of classes and complex settings, emphasizing small-sized objects. We also proposed an RFA-based MCGR network that achieved state-of-the-art image SR quality and object detection tasks. The current detection accuracy for the classes like vehicle, airplane, and ship are satisfactory, while there needs further exploration to learn the complex features of the tree and low-vegetation classes.

Extensive experiments show that using an image SR network before the object detection task helps in improving the mAP for object detection, and the proposed MCGR outperforms the state-of-the-art YOLOv5 for mAP by 5% and 13% for scale factors of 2 and 4, respectively. Furthermore, we also shared the results of object detection on an independent dataset which shows the flexibility of the proposed MCGR network to perform object detection on other datasets.

## CRediT Authorship Contribution Statement

**Yi Wang**: Conceptualization, funding acquisition, formal analysis, methodology, supervision of final version, review and editing.

**Syed Muhammad Arsalan Bashir**: Conceptualization, formal analysis, investigation of results, methodology, performing experiments on the state-of-the-art methods, collection and annotation of the dataset, development of the MCGR pipeline and training and validation of results, writing - original draft, writing – final version, review and editing.

**Mahrukh Khan**: Conceptualization, Performing experiments on the state-of-the-art methods, Collection and annotation of the dataset, development of the MCGR pipeline and training and validation of results.

**Qudrat Ullah, Rui Wang** and **Yilin Song**: Collection and annotation of the dataset, running experiments in current image SR methods.

**Zhe Guo** and **Yilong Niu**: Help in Analysis and Overall supervision.

## Acknowledgements

This work is supported by the National Natural Science Foundation of China under Grant no. 62071384.